\def\year{2019}\relax
\documentclass[letterpaper]{article}

\usepackage{aaai}
\usepackage{times}
\usepackage{helvet}
\usepackage{courier}
\usepackage{amssymb}
\usepackage{amsmath}
\usepackage{amsthm}
\usepackage{color}
\usepackage[utf8]{inputenc}
\usepackage{graphicx}
\usepackage{dsfont}
\usepackage{algorithmicx}
\usepackage{algpseudocode}
\usepackage{xspace}
\usepackage{natbib}
\usepackage{hhline}
\usepackage[hyphens]{url}
\usepackage{comment}
\usepackage[makeroom]{cancel}

\usepackage{subcaption}

\usepackage[compact]{titlesec}         
\titlespacing{\section}{0pt}{5pt}{0pt} 

\title{The Actor-Advisor: Policy Gradient With Off-Policy Advice}

\author{
	H\'el\`ene Plisnier,\textsuperscript{1}
	Denis Steckelmacher,\textsuperscript{1}
	Diederik M. Roijers,\textsuperscript{2}
	Ann Now\'e\textsuperscript{1} \\
	\textsuperscript{1} Vrije Universiteit Brussel \\
	\textsuperscript{2} Vrije Universiteit Amsterdam
}

\allowdisplaybreaks
\begin{document}
\maketitle

\begin{abstract}


Actor-critic algorithms learn an explicit policy (\emph{actor}), and an accompanying value function (\emph{critic}). The actor performs actions in the environment, while the critic evaluates the actor's current policy. However, despite their stability and promising convergence properties, current actor-critic algorithms do not outperform critic-only ones in practice. We believe that the fact that the critic learns $Q^{\pi}$, instead of the optimal Q-function $Q^*$, prevents state-of-the-art robust and sample-efficient off-policy learning algorithms from being used. In this paper, we propose an elegant solution, the Actor-Advisor architecture, in which a Policy Gradient actor learns from unbiased Monte-Carlo returns, while being shaped (or \emph{advised}) by the Softmax policy arising from an off-policy critic. The critic can be learned independently from the actor, using any state-of-the-art algorithm. Being advised by a high-quality critic, the actor quickly and robustly learns the task, while its use of the Monte-Carlo return helps overcome any bias the critic may have. In addition to a new Actor-Critic formulation, the Actor-Advisor, a method that allows an external advisory policy to shape a Policy Gradient actor, can be applied to many other domains. By varying the source of advice, we demonstrate the wide applicability of the Actor-Advisor to three other important subfields of RL: safe RL with backup policies, efficient leverage of domain knowledge, and transfer learning in RL. Our experimental results demonstrate the benefits of the Actor-Advisor compared to state-of-the-art actor-critic methods, illustrate its applicability to the three other application scenarios listed above, and show that many important challenges of RL can now be solved using a single elegant solution.

\end{abstract}

\section{Introduction}
Actor-critic algorithms \citep{Barto1983,Konda1999} learn an explicit actor policy that executes actions in the environment, while the critic evaluates the actions selected by the actor. Despite promising stability and convergence guarantees, actor-critic methods \citep{Wang2016b,Gruslys2017,Schulman2017} are still outperformed by critic-only ones \citep{Schaul2015,Anschel2017,Medina2018}. In our opinion, the weakness of actor-critic algorithms lies in their critic being \emph{on-actor}, i.e., it learns $Q^{\pi}$, the function that evaluates the current actor $\pi$, instead of $Q^*$, the optimal Q-function. Producing accurate estimates of $Q^{\pi}$ is difficult, and limits the extent to which experience replay can be used \citep{Wang2016b}, which reduces sample-efficiency.

In this paper, we present the Actor-Advisor, an architecture in which a Policy Gradient actor learns from its Monte-Carlo returns, instead of Q-Values provided by its critic, and a Double DQN \citep{Hasselt2016} critic advises the actor before an action is selected. To let the critic advise the actor, Policy Shaping \citep{Fernandez2006,Griffith2013} is used. Policy Shaping allows an agent's policy to be directly influenced by an external advisory policy by sampling actions from their mixture. Until now, Policy Shaping could not be used with Policy Gradient, since the action being executed by a vanilla Policy Gradient agent has to be drawn from the exact probability distribution output by the policy. Otherwise, learning diverges \cite{Sutton2000}. Our contribution (Section \ref{sec:contribution}) addresses this convergence issue, and allows the Softmax policy arising from the critic to be mixed with the actor's policy, without impairing convergence.

We compare our contribution, the Actor-Advisor, to state-of-the-art actor-critic algorithms available in the OpenAI baselines, such as Proximal Policy Optimization \citep{Schulman2017} and A3C \citep{Mnih2016}. We also compare the Actor-Advisor, which components are Policy Gradient and Double DQN, with its critic-only counterpart, i.e., Double DQN. Although it is not current state-of-the-art, Double DQN is readily available, easy to implement and tune, and largely outperforms state-of-the-art actor-critic algorithms. Our results demonstrate that the Actor-Advisor's performance is much superior to conventional actor-critic algorithms', and shows better sample-efficiency than critic-only Double DQN \citep{Hasselt2016}. We further note that our Actor-Advisor algorithmic architecture is compatible with any type of critic. In addition to advice emanating from a sample-efficient off-policy critic, we also demonstrate the applicability of the Actor-Advisor to other settings, where advice comes from various sources: a safety-critical task in which the advisor is a proven-safe backup policy; a complex tasks for which a simple and sub-optimal heuristic policy is available; and a transfer learning setting in which a previously learned policy in a given environment advises the agent in a similar but distinct environment.

\section{Background}
\label{sec:background}

In this section, we formally introduce the concepts at the basis of Actor-Advisor, that is Markov Decision Processes (MDPs), value-based methods, Policy Gradient compared to actor-critic methods and how policies in the form of probability distributions are combined to allow Policy Shaping. 

\subsection{Markov Decision Processes}
\label{sec:background_mdp}

A discrete-time Markov Decision Process (MDP) \citep{Bellman1957} with discrete actions is defined by the tuple $\langle S, A, R, T \rangle$: a possibly-infinite set $S$ of states; a finite set $A$ of actions; a reward function $R(s_t, a_t, s_{t+1}) \in  \mathbb{R}$ returning a scalar reward $r_t$ for each state transition; and a transition function $T(s_{t+1} | s_t, a_t) \in [0, 1]$ taking as input a state-action pair $(s_t, a_t)$ and returning a probability distribution over new states $s_{t+1}$.

A stochastic stationary policy $\pi(a_t | s_t) \in [0, 1]$ maps each state to a probability distribution over actions. At each time-step, the agent observes $s_t$, selects $a_t \sim \pi(s_t)$, then observes $r_{t+1}$ and $s_{t+1}$. The $(s_t, a_t, r_{t+1}, s_{t+1})$ tuple is called an \emph{experience} tuple. An optimal policy $\pi^*$ maximizes the expected cumulative discounted reward $E_{\pi^*}[\sum_t \gamma^t r_t]$, where $\gamma$ is a discount factor. The goal of the agent is to find $\pi^*$ based on its experiences within the environment.

\subsection{Value-based methods}
\label{sec:background_value-based}

Q-Learning \citep{Watkins1989} learns the optimal action-value function $Q^* (s, a) = E_{\pi^*} [\sum_{k=0}^{\infty} \gamma^k r_{t+k} | s_t = s, a_t = a]$, i.e., the optimal expected return for each action in each state. Value-based methods, such as Q-Learning and SARSA \citep{Rummery1994}, select actions according to the Q-Values and some exploration strategy. Then, the resulting experience tuple is used to update the Q-function according to this formula:

\begin{align}
    \label{eq:qlearning}
    Q_{k+1}(s_t, a_t) &= Q_k (s_t, a_t) + \alpha \delta, \\
    \nonumber
    \delta &= r_{t+1} + \gamma \max_a Q_k (s_{t+1}, a) - Q_k (s_t, a_t)
\end{align}
The SARSA update rule is obtained by replacing $\max_{a} Q_k(s_{t+1}, a)$ with $Q_k(s_{t+1}, a_{t+1})$, where $a_{t+1}$ is the action actually performed by the agent in the environment. This changes the Q-function being learned, that becomes $Q^{\pi}$ instead of $Q^*$, with $\pi$ the policy that represents the actual behavior of the agent.

\subsection{Policy Gradient and Actor-Critic Algorithms}
\label{sec:background_pg}
Instead of choosing actions according to Q-Values, Policy Gradient methods \citep{Williams1992,Sutton2000} explicitly learn an \emph{actor} policy $\pi_\theta(a_t | s_t) \in [0, 1]$, parametrized by a weights vector $\theta$, such as the weights of a neural network. The objective of the agent is to maximize the expected cumulative discounted reward $E_{\pi}[\sum_t \gamma^t r_t]$, which translates to the minimization of the following equation \citep{Sutton2000}:

\begin{align}
	\label{eq:pg}
	\mathcal{L}(\pi_{\theta}) &= -\sum\limits_{t=0}^{T} \left\{
	\begin{array}{cl}
		\mathcal{R}_t \\
		Q^{\pi_{\theta}}(s_t, a_t)
	\end{array}
	\right\} \log (\pi_{\theta} (a_t | s_t))
\end{align}

\noindent
with $a_t \sim \pi_{\theta}(s_t)$ the action executed at time $t$, $\mathcal{R}_t = \sum_{\tau=t}^{T} \gamma^{\tau} r_{\tau}$ the Monte-Carlo return at time $t$, and $r_{\tau} = R(s_{\tau}, a_{\tau}, s_{\tau+1})$ the immediate reward. At every training epoch, experiences are used to compute the gradient $\frac{\partial \mathcal{L}}{\partial \theta}$ of Equation \ref{eq:pg}, then the weights of the policy are adjusted one small step in the opposite direction of the gradient. A second gradient update requires fresh experiences \citep{Sutton2000}, which prevents Policy Gradient to be used with experience replay to improve sample-efficiency. 

Actor-critic methods allow the actor to use Q-Values instead of Monte-Carlo returns (the returns $\mathcal{R}_t$ are replaced by $Q^{\pi_{\theta}}(s_t, a_t)$ in Equation \ref{eq:pg}), which leads to a gradient of lower variance. Such methods achieve impressive results on several challenging tasks \citep{Wang2016b,Gruslys2017,Mnih2016,Schulman2017}. However, actor-critic algorithms rely on their critic being on-actor, i.e., the learned $Q^{\pi_{\theta}}$ must be accurate for the current actor. Maintaining a close and healthy relationship between the actor and the critic requires careful design, as discussed by \cite{Konda1999} and \cite{Sutton2000}.

\subsection{On-Policy, Off-Policy and On-Actor}

The definitions of on-policy, off-policy and \emph{on-actor} are central concepts in this paper. Their definitions are as follows. Off-policy algorithms learn the Q-function of a policy that is not the one being executed by the agent. For instance, Q-Learning learns the optimal Q-function $Q^* \equiv Q^{\pi*}$ while the agent executes another policy $\pi$. On the contrary, on-policy algorithms, like SARSA, learn the Q-function $Q^{\pi}$ that evaluates the executed policy $\pi$. While generally accepted, this definition is sometimes unclear when it comes to describing the critic being learned by actor-critic algorithms, since two policies intervene in this setting: the actor $\pi$, and the behavior policy $\mu$. While $\pi = \mu$ in most cases, recent work allows $\mu$ to be distinct from $\pi$ \citep{Degris2012,Wang2016b,Gu2017b}. We therefore denote as \emph{on-actor} an algorithm that learns $Q^{\pi}$ evaluating the actor's policy $\pi$ (regardless of what $\mu$ is), and \emph{off-actor} an algorithm that learns $Q^*$.

\subsection{Policy Shaping}
\label{sec:background_mixing}

Policy Shaping \citep{Griffith2013} allows the agent's learned policy $\pi^L(s_t)$ to be influenced by an external advisory policy $\pi^E(s_t)$. The agent samples actions from the mixture of the two policies $\pi^L(s_t)$ and $\pi^E(s_t)$, computed by performing an element-wise multiplication of $\pi^L(s_t)$ with $\pi^E(s_t)$, divided by their dot product, as follows:

\begin{align}
    \label{eq:pss}
    a_t \sim  \frac{\pi^L(s_t) \pi^E(s_t)} {\pi^L(s_t) \cdot \pi^E(s_t)}
\end{align}

\noindent
Even if first proposed for integrating human feedback in the learning process \citep{Griffith2013}, this simple method can be applied to a larger variety of problems.

\section{The Actor-Advisor}
\label{sec:contribution}

Policy Gradient requires the actions taken by the agent to be directly sampled from the policy $\pi$ it is learning \citep{Sutton2000}, which prevents any exploration mechanism (e.g., $\varepsilon$-Greedy), backup, heuristic or advisory policy to be leveraged by the agent. Our main contribution, the Actor-Advisor architecture, consists of a solution to this problem, that allows the policy executed by the agent to be directly influenced by an external advisory policy, without impairing convergence. We first present the foundation of our contribution, then analyze its theoretical properties, and finally discuss the various ways it can be applied to real-world problems.

\subsection{Policy Gradient with External Advice}

The Actor-Advisor assumes a parametric policy represented by a neural network, trained using Policy Gradient \citep{Sutton2000}. The neural network has two inputs: the state $s_t$ and an \emph{advice} $\pi^E$, a probability distribution over actions. The state is used to compute a probability distribution over actions $\pi^L_{\theta}$, using dense, convolutional or any other kind of \textit{trainable} layers. Then, the output $\pi_{\theta}$ of the network is computed by element-wise multiplying $\pi^L_{\theta}$ and $\pi^E$, followed by a normalization \citep{Griffith2013}:

\vspace{-4mm}
\begin{align}
	\nonumber
    \pi_{\theta}(s_t, \pi^E(s_t)) &= \frac{\pi^L_{\theta}(s_t) \pi^E(s_t)} {\sum_{a \in A} \pi^L_{\theta}(a|s_t) \pi^E(a|s_t)} \\
    \label{eq:pss2}
   	&= \frac{\pi^L_{\theta}(s_t) \pi^E(s_t)} {\pi^L_{\theta}(s_t) \cdot \pi^E(s_t)}
\end{align}

\noindent where $\pi^L_{\theta}(s_t) \cdot \pi^E(s_t)$ is the dot product between the two vectors, and $\pi_{\theta}(s_t, \pi^E(s_t))$ denotes the parametric policy $\pi_{\theta}$ from which the action $a_t$ is actually sampled. The key insight is that the gradient is calculated based on the mixed policy $\pi_{\theta}(s_t, \pi^E(s_t))$, rather than just on the learned parametrized policy $\pi^L_{\theta}$. As a result, the distribution $\pi_{\theta}$ output by the network is a mixture of $\pi^L_{\theta}$ and some advice, and actions can therefore be directly sampled from that distribution. The element-wise multiplication, without any square root or other kind of transformation, has the nice property that an all-ones $\pi^E$ has a neutral effect on $\pi^L_{\theta}$, therefore disabling advice in an elegant way. The network is trained using the standard Policy Gradient loss \citep{Sutton2000}:

\begin{align}
    \label{eq:pgg}
    \mathcal{L}(\pi) &= -\sum\limits_{t=0}^{T} \mathcal{R}_t \log (\pi_{\theta}(a_t | s_t, \pi^E(s_t)))
\end{align}

\noindent with $\pi_{\theta}(a_t | s_t, \pi^E(s_t))$ the probability to execute action $a_t$ at time $t$, given as input the state $s_t$ and some state-dependent advice $\pi^E(s_t)$, and the return $\mathcal{R}_t = \sum_{\tau=t}^{T} \gamma^{\tau} r_{\tau}$, with $r_\tau = \mathcal{R}(s_\tau, a_\tau, s_{\tau+1})$, a simple discounted sum of future rewards. Note that the network training (or optimization) problem is not made harder by the use of $\pi^E$, since $\pi^E$ is not fed to the trainable part of the network, is not parametric, and as such cannot be considered as an extension of the state-space. This neural architecture, inspired by how variable action-spaces are implemented in \cite{Steckelmacher2018}, meets all the Policy Gradient assumptions, yet the behavior of the agent can be directly altered by an external advisory policy from any source. Moreover, our experimental results in Section \ref{sec:experiments} demonstrate that Actor-Advisor is able to leverage advice to learn faster.

\subsection{Stochastic and Deterministic Advice}
\label{subsec:contribution_StocDet}
Depending on the values of $\pi^E(a | s)$, two distinct forms of advice (in this section, we consider that almost deterministic advice is still stochastic) arise, leading to distinct potential uses of the advice:

\begin{description}
\item[Stochastic] advice ensures that each action has a non-zero probability, and can thus be selected by the agent with an arbitrarily high or low probability, depending on $\pi^L$. Stochastic advice can be seen as an optional, \emph{soft} advice, as it will merely bias the action selection rather than determining it. Stochastic advice can be used to allow a heuristic, a human teacher, or a previously learned policy (in a transfer learning setting) to help the agent learn the task. In the case of our new Actor-Critic formulation, the Actor-Advisor, in which a Policy Gradient actor is advised by a Double DQN critic, the advisory policy arising from the critic is also stochastic. In the above-mentioned cases, stochastic advice can be suboptimal, hence it is desirable that the agent learns to ignore it whenever following the advice endangers its performance. A property we believe inherent to Policy Gradient (on which the Actor-Advisor is based) is that the higher the entropy of the advice, the easier it is for the agent to learn to ignore it. Our experiments in Section \ref{sec:experiments} empirically confirms this intuition.
\item[Deterministic] advice has one 1 for a particular action, and a zero probability for all other actions. We consider that deterministic advice, or \emph{directives}, are used to enforce proven-safe mandatory guidelines that must be respected by the agent, in a safe RL task.
\end{description}

\subsection{Analyzing the Gradient}

Conventional Policy Gradient learns a policy that directly maps states to actions, as do our $\pi^L_{\theta}$ policy. In contrast to vanilla Policy Gradient, the gradient computed by Actor-Advisor is not only based on the policy learned $\pi^L_{\theta}$, but on a mixture of $\pi^L_{\theta}$ with the advice $\pi^E$, which is then normalized. In this section, we derive the gradient flowing into $\pi^L_{\theta}$, show that it is modified by $\pi^E$, and detail how $\pi^E$ influences the gradient.

We consider two cases: when the advice is deterministic, we show that the gradient flowing in $\pi^L_{\theta}$ is zero. When the advice is uniform (so, no actual advice is given), we show that the gradient flowing in $\pi^L_{\theta}$ is exactly the vanilla Policy Gradient. The general case, when the advice is stochastic but non-uniform, cannot be compared to vanilla Policy Gradient in our opinion: the Actor-Advisor gradient is different from the vanilla Policy Gradient, which is fortunate because vanilla Policy Gradient is not compatible with external advice. Nevertheless, our empiric results suggest that the Actor-Advisor finds a local optimum (just as vanilla Policy Gradient does), even if stochastic advice may change the path towards it. An in-depth analysis, that considers the whole learning dynamics of the agent instead of point estimates of the gradient, is beyond the scope of this paper.

The first step of our analysis consists of plugging Equation \ref{eq:pss2} in Equation \ref{eq:pgg}, which produces the general Actor-Advisor gradient. For clarity, we omit $s_t$ in our notations, which makes $\pi_{\theta}(a_t | s_t)$ read as $\pi_{\theta}(a_t)$:

\begin{align*}
    \nabla_{\theta} \mathcal{L}(\pi_{\theta}) &= -\nabla_{\theta} \sum\limits_{t=0}^{T} \mathcal{R}_t \log (\pi_{\theta}(a_t | s_t, \pi^E(s_t))) \\
    &= -\nabla_{\theta} \sum\limits_{t=0}^{T} \mathcal{R}_t \log \left( \frac{\pi^L_{\theta}(a_t) \pi^E(a_t)} {\pi^L_{\theta} \cdot \pi^E} \right) \\
    &= \begin{array}{ll}
    	-\nabla_{\theta} \sum\limits_{t=0}^{T} \mathcal{R}_t (\log(\pi^L_{\theta}(a_t)) &\cancel{+ \log(\pi^E(a_t))} \\
    	&- \log(\pi^L_{\theta} \cdot \pi^E))
    \end{array}
\end{align*}

We now consider two cases: i) the advice $\pi^E$ is deterministic; and ii) $\pi^E$ is uniform.

\paragraph{Deterministic} The advice $\pi^E$ is a deterministic vector with a single one, and the other actions have zeros; in other words, $a_t = \operatorname{argmax}_a \pi^E$, the only action allowed by $\pi^E$, and $\pi^E(a_t) = 1$. For compactness, we denote as $\nabla_{\theta} \mathcal{L}(\pi_{\theta})(t)$ the loss at time-step $t$, with $\nabla_{\theta} \mathcal{L}(\pi_{\theta}) = \sum_{t=0}^{T} \nabla_{\theta} \mathcal{L}(\pi_{\theta})(t)$:

\begin{align*}
	\nabla_{\theta} \mathcal{L}(\pi_{\theta})(t) &= -\nabla_{\theta} \mathcal{R}_t (\log(\pi^L_{\theta}(a_t)) - \log(\pi^L_{\theta}(a_t) 1)) \\
	&= -\nabla_{\theta} ~ 0
\end{align*}

In states where $\pi^E$ is deterministic, the gradient becomes zero, which prevents $\pi^L_{\theta}$ from changing. This is expected, as in those states, $\pi^E$ fully determines the actions being executed, and $\pi^L_{\theta}$ is ignored. This property is important, as it ensures that the parameters $\theta$ do not \emph{drift} when deterministic advice is used. Covering big parts of the environment with a backup policy is therefore safe, and does not impair learning in the other parts of the environment.

\paragraph{Uniform} We now consider the case where the advice $\pi^E$ is uniform, be it normalized or not, such that $\forall a: \pi^E(a) = P$. We assume that $\pi^L_{\theta}$ is normalized, such that $\sum_a \pi^L_{\theta}(a) = 1$ for any $\theta$.

\begin{align*}
	\nabla_{\theta} \mathcal{L}(\pi_{\theta})(t) &= -\nabla_{\theta} \mathcal{R}_t (\log(\pi^L_{\theta}(a_t)) - \log(\sum_a P \pi^L_{\theta}(a))) \\
	&= -\nabla_{\theta} \mathcal{R}_t (\log(\pi^L_{\theta}(a_t)) - \cancel{\log(P)}) \\
    &= -\nabla_{\theta} \mathcal{R}_t (\log(\pi^L_{\theta}(a_t)))
\end{align*}

In this case, $\log(P)$ does not depend on $\pi^L_{\theta}$, and can thus be removed from the gradient. We also note that, because  $\pi^L_{\theta}$ is normalized, the gradient of its norm is zero. When uniform advice is used, only the standard Policy Gradient remains, and the agent learns as if our contribution was not implemented.

This concludes our analysis of the gradient when uniform or deterministic advice is used. Our experimental results in Section \ref{sec:experiments} demonstrate the robustness of the Actor-Advisor to all forms of advice, even stochastic non-uniform one, which complements the analysis of this section.

\section{Application Domains}
\label{sec:review}

In this section, we review three subfields of reinforcement learning to which our main contribution, presented in Section \ref{sec:contribution}, can be applied. The three sub-fields are safe RL, efficient leverage of domain knowledge, and transfer learning.

\subsection{Safe RL}

A reinforcement learning agent is considered \emph{safe} if it does not behave in an undesirable and harmful manner \citep{Amodei2016}; in other words, it will not take actions leading to undesirable transitions (with ``undesirable" defined by some designer-provided constraints). According to \cite{Garcia2015} and \cite{Amodei2016}, to tackle the safe RL problem, we can either change the agent's optimization criterion, or, what we focus on in this paper, change its action selection strategy, possibly using a safe backup policy \citep{Hans2008}. Backup policies, regarded by engineers as easy to design, forcibly take control of the agent in certain circumstances, and move it to safe regions of the state-space. In our experiments, we consider a safety-critical task in which a robot navigates on a table, and is prevented from falling by a designer-provided backup policy (Section \ref{subsec:envs}). When the robot is too close to the edge of the table, the backup policy forces the robot to turn until it does not face the edge of the table any more.

\subsection{Leveraging Domain Knowledge}

Our use of advice through Policy Shaping is similar to dynamic Reward Shaping \citep{Ng1999}, which allows an expert designer, or human teacher, to help an RL agent learn faster in environments where rewards are sparse, by giving it additional rewards.
However, Reward Shaping only allows to express feedback \citep{Thomaz2006,Knox2009,Christiano2017,Pilarski2017}, i.e., evaluations of the actions performed by the agent \emph{after} their execution. Hence, in the case where the agent performs an undesirable action, it is only possible to complain a posteriori, and there is no way to prevent the agent from wrong doing.
Policy Shaping has been used to let a human critique policy, extracted from human-delivered feedback, influence the agent's actions by mixing it with the agent's policy \citep{Griffith2013}. However, because the critique policy is slowly learned from feedback, the undesirable action prevention problem remains. In our experiments, we instead consider \emph{advice} \citep{Clouse1992,Maclin1996,Thomaz2008}, provided by a heuristic, rather than feedback. In our setting, at every time-step, our heuristic can advise the agent on which action to choose before the agent has selected any action (see Section \ref{subsec:safe}). In contrast to using Reward Shaping and feedback, advising the agent through Policy Shaping makes it possible to halt or deflect the agent's behavior \emph{before} it engages in undesirable actions, by expressing a directive, i.e., deterministic advice (Section \ref{subsec:contribution_StocDet}).

\subsection{Transfer Learning}

From an RL perspective, transfer learning allows to generalize across tasks, and to reuse a policy learned in a given environment in a new similar environment \citep{Taylor2009}. \cite{Fernandez2006} propose an algorithm inspired by Policy Shaping, that, at each timestep, reuses with probability $\psi$ a previously learned policy, and greedily exploits the currently learned policy with probability $1-\psi$. Our contribution, the Actor-Advisor, can be applied in a transfer learning setting, and has a significant advantage on the algorithm of \citeauthor{Fernandez2006}: it does not require the $\psi$ hyperparameter to be defined, the mixing happens automatically.

Distillation is closely related to transfer learning. A large classifier generates samples used to train a smaller one, in the hope that the smaller classifier will be as good as the big one, but much more compact \citep{Bucila2006}. We show in Section \ref{sec:experiments_transfer} that Actor-Advisor naturally implements a robust distillation scheme. An external policy is distilled in the current one, even if the task these two policies solve is different (but on the same action set). The Actor-Advisor makes no assumption about where the \textit{old} policy comes from, or how the \textit{new} policy is represented, and can therefore be used to distil a policy to a much simpler encoding.


\section{Experiments}
\label{sec:experiments}

Our main contribution, the Actor-Advisor architecture, allows an external policy to guide and influence the policy being learned by the agent. This section illustrates how this general framework naturally applies to the following areas:

\begin{enumerate}
  \item \textbf{actor-critic:} We introduce a combination of Policy Gradient with Double DQN, made possible by our contribution, allowing Policy Gradient to be advised by the critic, instead of directly using the critic's Q-Values (Section \ref{subsec:dqn+pg}).
  \item \textbf{safety \& domain knowledge:} We evaluate the Actor-Advisor on a safety-critical task, in which a safe backup policy prevents the agent from executing a dangerous action such as falling off a table, and some additional heuristic advice helps the agent when it is close to the goal (Section \ref{subsec:safe}).
  \item \textbf{transfer learning:} We illustrate how an agent, that has to navigate in a house, is able to leverage advice from a policy learned in another house to learn faster (Section \ref{sec:experiments_transfer}).

\end{enumerate}

\subsection{Environments}
\label{subsec:envs}

Our experiments take place in two environments which we introduce below. Table has continuous states, complex dynamics and is difficult to explore. Five Rooms is a smaller environment, but that leads to challenging transfer learning tasks.

\begin{figure}
	\centering
	\includegraphics[width=45mm]{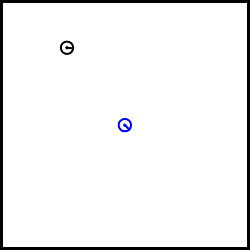}
	\caption{The Table environment: the agent (in black) must align itself to the charging station (in blue) to plug itself to it, without falling from the table by getting too close to the edges.}
    \label{fig:table}
\end{figure}

\noindent \textbf{Table} is a continuous-states and discrete-actions environment in which an agent must find and plug itself to a charging station, resulting in a +100 reward, without getting too close to the edges and fall off the table, which results in a -50 reward. The table is a one-by-one unit square. Three actions are available to the agent: go forward 0.005 units, turn left 0.1 radians, and turn right 0.1 radians. The agent observes its $(x, y)$ position and its current orientation $\theta$, expressed in radians. It starts in position $(0.1, 0.1)$ with $\theta=0.1$. The episode terminates either if the agent successfully docks itself on the charging station, i.e., if its position is $(0.5 \pm 0.05, 0.5 \pm 0.05)$ and $\theta=\frac{\pi}{4} \pm 0.3$, or after 2000 unfruitful timesteps, or if the agent falls off the table. Because the agent moves slowly compared to the size of the table, and that the reward signal is sparse, this task is extremely challenging. It is more difficult to explore than most Gym tasks, and possibly comparable in difficulty to some Atari games.

\begin{figure}
	\centering
	\includegraphics{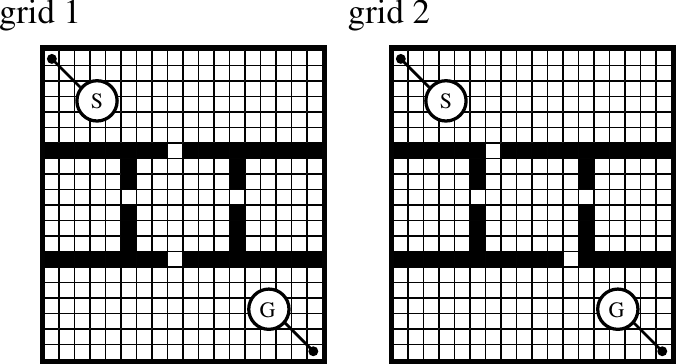}
	\caption{The two $20 \times 18$ Five Rooms grids, ideal for transfer learning tasks. Black cells represent walls; the agent starts in the initial state S, and must reach the goal G. The difference between grid 1 and grid 2 is the location of the two doors on the path from S to G.}
	\label{fig:grids}
\end{figure}

\noindent \textbf{Five Rooms} is a $20 \times 18$ cells grid world environment (Figure \ref{fig:grids}). It is divided into five rooms, and each room is accessible via one of the four one-cell-wide doors. This, in addition to its size, makes exploration difficult. The agent can move one cell up, down, left or right, unless the target cell is a wall (then the agent does not move). The agent starts in the top-left corner of the grid, and must reach the bottom right corner, where it receives a reward of $+100$; it receives $-1$ in every other cell. The episode terminates either once the goal has been reached, or after 500 unfruitful time-steps. The optimal policy takes $35$ time-steps to reach the goal, and obtains a cumulative reward of $65$. For the transfer learning experiment, we use two slightly different variants of the same grid: in grid 1, the two main doors are centred; in grid 2, the doors are shifted.

\subsection{Policy Gradient Advised by Double DQN}
\label{subsec:dqn+pg}

\begin{figure}[t]
	\centering
	\includegraphics{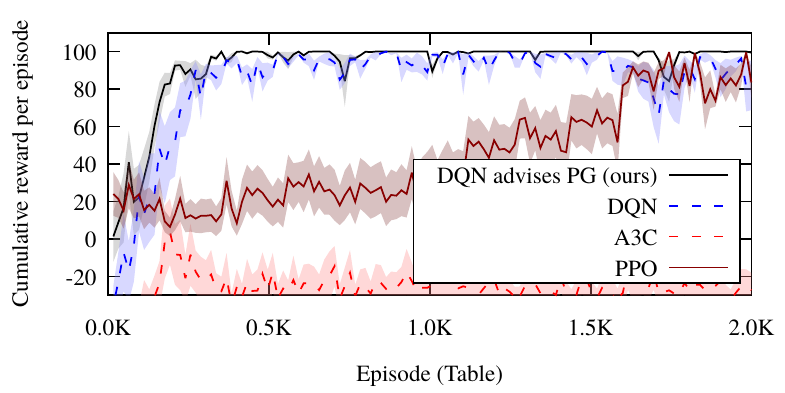}
	\includegraphics{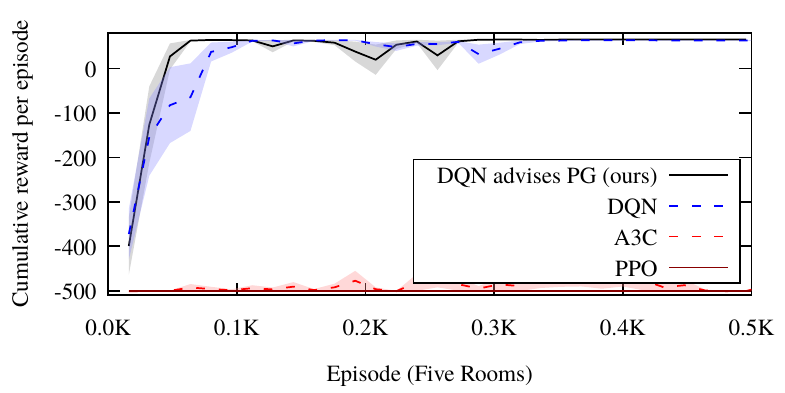}
	\caption{Our Actor-Advisor agent, in which the Policy Gradient actor is advised by the Softmax policy of a Double DQN critic, is much more sample-efficient than A3C, PPO and ACER (ACER learns after 10K episodes, so has been omitted from the plot), and outperforms the Double DQN critic used without an actor.}
	\label{fig:dqn+pg}
\end{figure}

We propose Actor-Advisor, an alternative to Actor-Critic, in which a Double DQN \citep{Hasselt2016} critic learns $Q^*$ from experiences, and provides advice to a Policy Gradient actor learning the same task. The Double DQN critic maintains an experience buffer of 20000 experiences, performs a training iteration every 16 time-steps, and replays 512 experiences per training iteration. This configuration, and intense experience replay, leads to stable and highly sample-efficient learning. At every timestep, the Softmax policy (with a temperature of 0.1) produced by the critic described above is mixed with the policy of the Policy Gradient agent. The Policy Gradient agent has a learning rate of 0.0001, and performs a gradient step every 16 episodes. Contrary to conventional Actor-Critic (Section \ref{sec:background_pg}), we keep the standard Policy Gradient loss unchanged (see Equation \ref{eq:pgg}), letting Policy Gradient learn from its returns instead of replacing them with Q-Values, that would then need to be on-policy. This addresses two important problems: it removes the bias introduced by potentially incorrect Q-Values, and it allows our highly sample-efficient off-policy critic to be used. The Actor-Advisor is much simpler than previous attempts at an actor-critic algorithm with experience replay \citep{Schulman2017,Mnih2016,Gruslys2017}, and achieves significantly higher sample-efficiencies.

We compare the Actor-Advisor against A3C \citep{Mnih2016}, ACER \cite{Wang2016b} and PPO \citep{Schulman2017}, three state-of-the-art actor-critic algorithms,\footnote{PPO optimizes an advanced policy loss, combined with a critic loss, which makes it actor-critic.} on the two environments described in Section \ref{subsec:envs}. In Figure \ref{fig:dqn+pg},\footnote{Results are averaged over 8 runs.} we show that the Actor-Advisor agent is as sample-efficient as plain Double DQN, and even displays a statistically significantly better sample-efficiency at the beginning of learning (p=1.9e-16 for Table and p=3.6e-26 for Four Rooms\footnote{We performed a Wilconxon test on 320 points in the [140, 180] episode interval for Table, and in the [40, 80] episode interval for Four Rooms.}). Additional results on the LunarLander environment with discrete actions \citep{Gym} confirm this, but were not added to this paper due to space constraints. The Actor-Advisor is, to our knowledge, the first actor-critic-like method that successfully exploits the sample-efficiency of an off-policy critic. Compared to A3C, ACER and PPO, the Actor-Advisor bridges the gap between actor-critic and critic-only algorithms, and we are confident that future improvements to the Actor-Advisor will enable it to compare more and more favorably to critic-only algorithms. In the next section, we evaluate the Actor-Advisor in other settings, where advice comes from sources other than a critic, and demonstrate its wide and successful applicability to various problems.


\subsection{Safe Backup Policy and Advice}
\label{subsec:safe}

\begin{figure}
	\centering
	\includegraphics{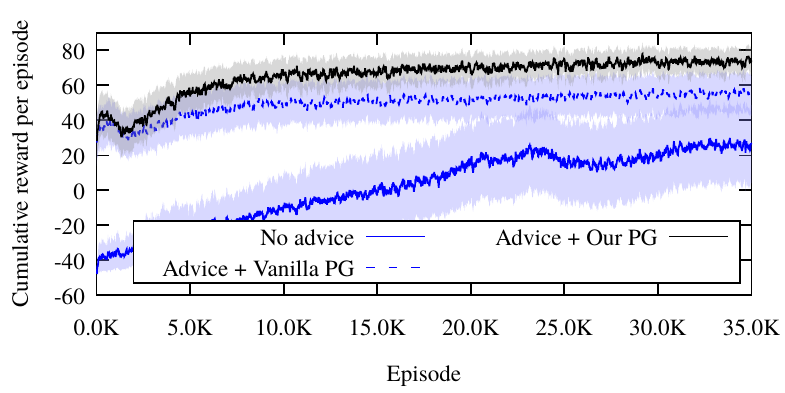}
	\caption{Reinforcement Learning with a backup policy and advice, on \emph{Table}: when advice and a backup policy are used, our Actor-Advisor (\textit{Advice + Our PG}) outperforms vanilla Policy Gradient (\textit{Advice + Vanilla PG}). The advisory and backup policies are crucial to the learning performance, and our Actor-Advisor architecture is able to leverage them to learn faster and more effectively.}
	\label{fig:backup}
\end{figure}


The Table environment is an example of a simulated robotic task in which a safe backup policy, as well as some heuristic advice, could be used by the agent. We consider: i) a designer-provided backup policy, that deflects the trajectory of the agent by forcing it to choose the turn action whenever it gets dangerously close to the edges of the table (the backup policy only kicks in when the agent is too close to an edge, and facing it); and ii) an advisory policy produced by a heuristic that, when the agent is close to the goal but in the incorrect orientation, makes it turn until the (successful) end of the episode. This advisory policy is simple and illustrates the benefits to be obtained from even the most basic heuristics. Both the safe backup policy and the heuristic advice are provided to the agent in the form of directives (Section \ref{subsec:contribution_StocDet}). When none of the heuristic or backup policies is activated, a neutral advice of all ones is given to it, thereby providing no direction at all.

In Figure \ref{fig:backup}, we compare our agent (\textit{Advice + Our PG}), advised by the safe backup and the heuristic policies, with vanilla Policy Gradient advised by the same policies (\textit{Advice + Vanilla PG}). In the \textit{Advice + Vanilla PG} setting, the actions of Policy Gradient are simply overridden by the external policies whenever they intervene, and the gradient is computed only based on the agent's learned policy (Section \ref{sec:contribution}). The \textit{No advice} setting learns the task slowly; the performance of the Actor-Advisor is superior to the naive \textit{Advice + Vanilla PG} setting. This demonstrates the necessity of our method to allow Policy Shaping to be used with Policy Gradient, while still allowing the optimal policy to be learned.

\subsection{Transfer Learning}
\label{sec:experiments_transfer}

Agents advising each other can be an effective way to transfer knowledge from one task to another. We evaluate the Actor-Advisor in a transfer learning setting, where an agent learns a policy to navigate in grid 1 (Five Rooms environment, Figure \ref{fig:grids}), policy that then advises a new agent learning in the second grid. This task illustrates the real-world setting of a robot that has to move in a house that changes over time, or that has been trained in one house and has to be moved to another one. We first train a conventional Policy Gradient agent on grid 1, and save its learned policy $\pi_1$ once its entropy almost reaches 0. Then, we launch our Actor-Advisor agent on grid 2, and let the policy $\pi_1$ advise our agent on grid 2. To make sure that our agent can ignore the advice from $\pi_1$ in the parts of the environment where it is irrelevant, we use $\pi^E(s, a) = \pi_1(s, a) + 1$, with the 1 ensuring that the entropy of the advice is high enough (the Actor-Advisor always normalizes the mixture of $\pi^E$ and $\pi^L$, so an unnormalized $\pi^E$ can be used as-is). In this configuration, the actor learns to find the goal in grid 2, while the advisor $\pi_1$ remains fixed ($\pi_1$ is not updated with new data from grid 2).

\begin{figure}
	\centering
	\includegraphics{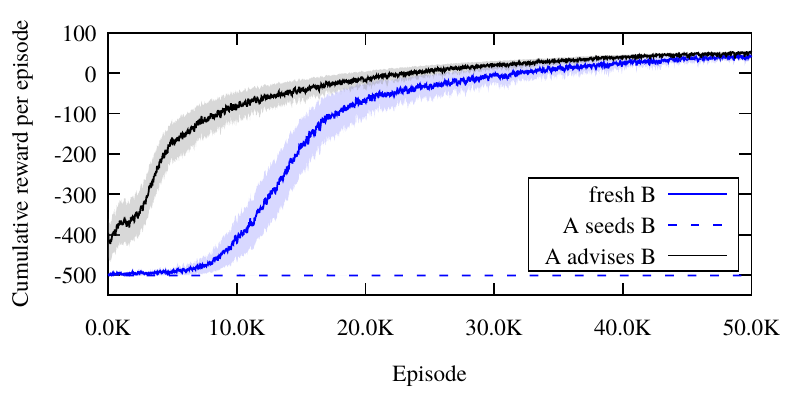}
	\caption{Transfer learning: three agents evaluated on grid 2: \textit{fresh B} is a fresh Policy Gradient agent; \textit{A seeds B} is a PG agent having learned to navigate in grid 1, and then launched as is on grid 2; \textit{A advises B} is our Actor-Advisor agent freshly launched on grid 2 while being advised by a policy learned on grid 1.}
	\label{fig:transfer}
\end{figure}

We compare our Actor-Advisor architecture to a more naive approach to transferring knowledge from grid 1 to grid 2, which is to simply relaunch the PG agent confident on grid 1 (and that has learned $\pi_1$), on grid 2, and let it try to re-adjust itself to its new environment. Surprisingly, an agent confident in some task shows great difficulty to re-adjust itself to a few changes in its environment. Our Actor-Advisor agent, on the other hand, shows robustness to the partially irrelevant stochastic advice given by its advisor $\pi_1$, and successfully leverages correct advice, resulting in a much faster learning than a new PG agent freshly launched on grid 2 (Figure \ref{fig:transfer}).

\section{Conclusion and Future Work}

Despite strong theoretical convergence guarantees and regret bounds, Actor-Critic suffers from its use of an on-actor critic that learns $Q_{\pi}$, the function evaluating the current actor $\pi$, instead of $Q^*$. This prevents actor-critic algorithms from using experience replay to improve sample-efficiency, and might be the reason why well-designed critic-only algorithms exhibit much higher sample-efficiency than actor-critic ones. We propose the Actor-Advisor, a novel combination of a Policy Gradient actor with a Q-Learning-based critic, in which the actor is advised by its critic before action selection, using Policy Shaping. We extend Policy Gradient so that it can be used with Policy Shaping without convergence issue. The Actor-Advisor allows the critic to fully leverage sample-efficient off-policy algorithms with experience replay, and to advise the actor about the optimal Q-function $Q^*$. We empirically show that the Actor-Advisor, combining Policy Gradient with Double DQN, outperforms well-known actor-critic algorithms, such as Proximal Policy Optimization and A3C, and even displays better sample-efficiency than Double DQN. To demonstrate the wide applicability of the Actor-Advisor, we evaluate it in three important RL subfields, by varying the source of advice: safe RL; efficient leverage of domain knowledge; and transfer learning. We believe that this work will lead to many more future research opportunities and applications.

\section*{Acknowledgements}

The first and second authors are funded by the Science Foundation of Flanders (FWO, Belgium), respectively as 1SA6619N Applied Researcher, and 1129319N Aspirant.

\bibliographystyle{aaai}
\bibliography{biblio}

\end{document}